\documentclass[twoside,11pt]{article}

\usepackage{jmlr2e}

\usepackage{hyperref}
\usepackage{mdwlist}
\usepackage{listings}
\usepackage[autolinebreaks,useliterate]{mcode}
\usepackage{algpseudocode}
\usepackage{algorithm}
\usepackage{amsmath}

\jmlrheading{1}{2013}{1-48}{4/00}{10/00}{Jakub Kone\v{c}n\'y and Michal Hagara}

\ShortHeadings{One-shot Gesture Recognition using HOG-HOF features}{Kone\v{c}n\'y and Hagara}
\firstpageno{1}

\begin{document}

\title{One-Shot-Learning Gesture Recognition using HOG-HOF Features}

\author{\name Jakub Kone\v{c}n\'y \email kubo.konecny@gmail.com \\
       \addr Comenius University\\
       Bratislava, Slovakia
       \AND
       \name Michal Hagara \email michal.hagara@gmail.com \\
       \addr Comenius University\\
       Bratislava, Slovakia}

\editor{Who knows}

\maketitle

\begin{abstract}
The purpose of this paper is to describe one-shot-learning gesture recognition systems developed on the \textit{ChaLearn Gesture Dataset} \citep{ChaLearn}. We use RGB and depth images and combine appearance (Histograms of Oriented Gradients) and motion descriptors (Histogram of Optical Flow) for parallel temporal segmentation and recognition. The Quadratic-Chi distance family is used to measure differences between histograms to capture cross-bin relationships. We also propose a new algorithm for trimming videos --- to remove all the unimportant frames from videos. We present two methods that use a combination of HOG-HOF descriptors together with variants of a Dynamic Time Warping technique. Both methods outperform other published methods and help narrow the gap between human performance and algorithms on this task. The code is publicly available in the MLOSS repository.
\end{abstract}



\section{Introduction}
\label{sec:introduction}

Gesture recognition can be seen as a way for computers to understand human body language. Thus, improving state-of-the-art algorithms for gesture recognition facilitates human-computer communication beyond primitive text user interfaces or GUIs (graphical user interfaces). With rapidly improving comprehension of human gestures we can start building NUIs (natural user interfaces) for controlling computers or robots. With the availability of such technologies, conventional input devices, such as a keyboard or mouse, could be replaced in situations in which they are inconvenient in future. Other applications of gesture recognition include sign language recognition, socially assistive robotics and game technology.

In this paper, we focus on the one-shot learning gesture recognition problem, in particular the \emph{ChaLearn Gesture Dataset} \citep{ChaLearn}. The dataset was released jointly with a competition, where the goal was to develop a system capable of learning to recognize new categories of gestures from a single training example of each gesture. The large dataset of hand and arm gestures was pre-recorded using an infrared sensor, Kinect$^{TM}$, providing both RGB and depth images \citep{Guyon, Guyon2}.

The purpose of this work is to describe methods developed during the \textit{ChaLearn Gesture Challenge} by the Turtle Tamers team (authors of this paper). We finished in 2$^{nd}$ place in round $2$ and were invited to present our solution at the International Conference on Pattern Recognition 2012, Tsukuba, Japan. The code has been made publicly available in the MLOSS repository.\footnote{\href{https://mloss.org/software/view/448}{https://mloss.org/software/view/448}}

Since the goal of the challenge was to provide solid baseline methods for this dataset, our methods were specifically tailored for this particular competition and dataset. Hence, they lack a certain generality, and we discuss and suggest changes for more general settings later.

The rest of this work is organised as follows. Related work is summarized in Section~\ref{sec:work}. In Section~\ref{sec:data} we describe the dataset and the problem in detail. In Section~\ref{sec:preprocessing} we focus on the preprocessing needed to overcome some of the problems in the dataset. Section~\ref{sec:feature} covers feature representation, using Histogram of Oriented Gradients and Histogram of Optical Flow, as well as a method used to compare similarities between these representations. In Section~\ref{sec:dtw} we describe the actual algorithms, and in Section~\ref{sec:results} we briefly describe algorithms of other participants and compare their results with ours, as well as with other published works. In Section~\ref{sec:conclusion} we summarize our paper and suggest an area for future work.


\section{Related  Work}
\label{sec:work}

In this section we provide a brief literature review in the area of gesture and action recognition and motivate our choices of models.

One possible approach to the problem of gesture recognition consists of analysing motion descriptors obtained from video. \citet{ikizler} use the output of Human Motion Capture systems in combination with Hidden Markov Models. \citet{Zhu} use Extended Motion History Image as a motion descriptor and apply the method to the \emph{ChaLearn Gesture Dataset}. They fuse dual modalities inherent in the Kinect sensor using Multiview Spectral Embedding \citep{MSE} in a physically meaningful manner.

A popular recent approach is to use Conditional Random Fields (CRF). \citet{wang06} introduce the discriminative hidden state approach, in which they combine the ability of CRFs to use long range depencencies and the ability of Hidden Markov Models to model latent structure. More recent work \citep{chatzis} describes joint segmentation and classification of sequences in the framework of CRFs. The method outperforms other popular related approaches with no sacrifices in terms of the imposed computational costs.

An evolution of Bag-of-Words \citep{bow}, a method used in document analysis, where each document is represented using the apparition frequency of each word in a dictionary, is one of the most popular in Computer Vision. In the image domain, these words become visual elements of a certain visual vocabulary. First, each image is decomposed into a large set of patches, obtaining a numeric descriptor. This can be done, for example, using SIFT \citep{lowe}, or SURF \citep{surf}. A set of $N$ representative visual words are selected by means of a clustering process over the descriptors in all images. Once the visual vocabulary is defined, each image can be represented by a global histogram containing the frequencies of visual words. Finally, this histogram can be used as input for any classification technique. Extensions to image sequences have been proposed, the most popular being Space-Time Interest Points \citep{laptev}. \citet{wang} have evaluated a number of feature descriptors and bag-of-features models for action recognition. This study concluded that different sampling strategies and feature descriptors were needed to achieve the best results on alternative action data sets. Recently an extension of these models to the RGB-D images, with a new depth descriptor was introduced by \citet{BoVDW}.

The methods outlined above usually ignore particular spatial position of a descriptor. We wanted to exploit the specifics of the dataset, particularly the fact that user position does not change within the same batch, thus also the important parts of the same gestures will occur roughly at the same place. We use a combination of appearance descriptor, Histogram of Oriented Gradients \citep{HOG} and local motion direction descriptor, Histogram of Optical Flow \citep{HOF}. We adopted Quadratic-Chi distance \citep{Qchi} to measure differences between these histograms. This approach only works well at high resolutions of descriptors. An alternative may be to use a non-linear support vector machine with a $\chi^2$ kernel \citep{laptev2}. Another possible feature descriptor that includes spatio-temporal position of features could be HOG3D \citep{klaser}, which was applied to this specific dataset by \citet{Fanello}.


\section{Data and Problem Setting}
\label{sec:data}

In this section, we discuss the easy and difficult aspects of the dataset and state the goal of the competition.

The purpose of the \emph{ChaLearn Gesture Challenge}\footnote{Details and website: \href{http://gesture.chalearn.org/}{http://gesture.chalearn.org/}.} was to develop an automated system capable of learning to recognize new categories of gestures from a single training example of each gesture. A large dataset of gestures was collected before the competition, which includes more than $50,000$ gestures recorded with the Kinect$^{TM}$ sensor, providing both RGB and depth videos. The resolution of these videos is $240 \times 320$ pixels, at 10 frames per second. The gestures are grouped into more than $500$ batches of $100$ gestures, each batch including $47$ sequences of $1$ to $5$ gestures drawn from small gesture vocabularies from $8$ to $14$ gestures. The gestures come from over $30$ different gesture vocabularies, and were performed by $20$ different users. 

During the challenge, development batches devel01-480 were available, with truth labels of gestures provided. Batches valid01-20 and final01-40 were provided with labels for only one example of each gesture class in each batch (training set). These batches were used for evaluation purposes. The goal is to automatically predict the gesture labels for the unlabelled gesture sequences (test set). The gesture vocabularies were selected from nine categories corresponding to various settings or applications, such as body language gestures, signals or pantomimes.


Easy aspects of the dataset include the use of a fixed camera and the availability of the depth data. Within each batch, there is a single user, only homogeneous recording conditions and a small vocabulary. In every sequence, different gestures are separated by the user returning to a resting position. Gestures are usually performed by hands and arms. In particular, we made use of the fact that the user is always at the same position within one batch.

The challenging aspects of the data are that within a single batch there is only one labelled example of each gesture. Between different batches there are variations in recording conditions, clothing, skin color and lightning. Some users are less skilled than others, thus there are some errors or omissions in performing the gestures. And in some batches, parts of the body may be occluded.

For the evaluation of results the Levenshtein distance was used, provided as the metric for the competition. That is the minimum number of edit operations (insertion, deletion or substitution) needed to be performed to go from one vector to another. For each unlabelled video, the distance $D(T,L)$ was computed, where $T$ is the truth vector of labels, and $L$ is our predicted vector of labels. This distance is also known as the ``edit distance''. For example, $D([1,2],[1]) = 1, \, D([1,2,3],[2,4]) = 2, \, D([1,2,3],[3,2]) = 2$.

The overall score for a batch was computed as a sum of Levenshtein distances divided by the total number of gestures performed in the batch. This is similar to an error rate (but can exceed 1). We multiply the result by a factor of $100$ to resemble the fail percentage. For simplicity, in the rest of this work, we call it the error rate.


\section{Preprocessing}
\label{sec:preprocessing}

In this Section we describe how we overcame some of the challenges with the given dataset as well as the solutions we propose. In Section~\ref{medianFilter} we focus on depth noise removal. Later we describe the need for trimming the videos --- removing set of frames --- and the method employed.

\subsection{Depth noise removal}
\label{medianFilter}

\begin{figure}[h!]
\centering
\includegraphics[width=\linewidth]{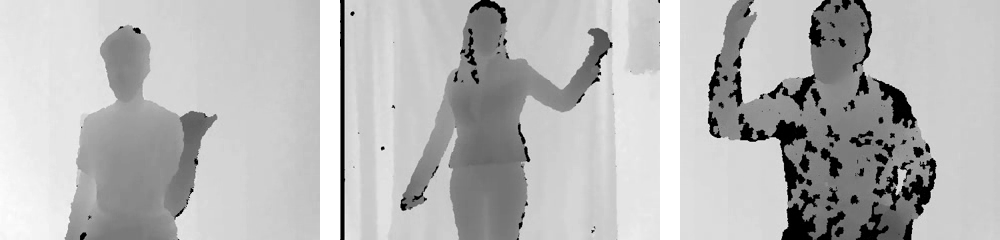}
\caption{Examples of depth images with various levels of noise}
\label{noise}
\end{figure}

One of the problems with the given dataset is the noise (or missing values) in the depth data. Whenever the Kinect sensor does not receive a response from a particular point, the sensor outputs a $0$, resulting in the black areas shown in Figure~\ref{noise}. This noise usually occurs along the edges of objects or, particularly in this dataset, humans. The noise is also visible if the object is out of the range of the sensor (0.8 to 3.5 meters).

The level of noise is usually the same within a single batch. However, there is a big difference in the noise level across different batches. If the level is not too high, it looks like `salt and pepper' noise. 

\begin{figure}[h!]
\centering
\includegraphics[width=\linewidth]{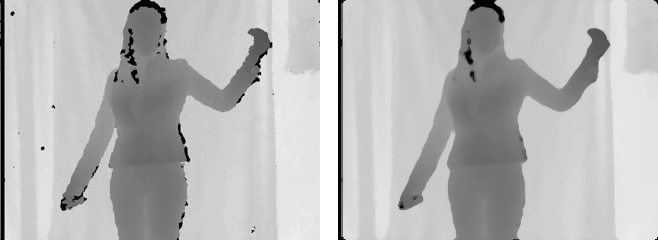}
\caption{Effect of median filter on depth image}
\label{medfilt}
\end{figure}

Later, in Section~\ref{sec:feature}, we use Histograms of Oriented Gradients (HOGs), which work best with sharp edges, so we need a filter that preserves the edges. One of the best filters for removing this kind of noise is the median filter, and also has our desired property. Median filter replaces every pixel with the median of pixels in small area around itself. The effect of the median filter is shown in Figure~\ref{medfilt}. We can see this filter does not erase big areas of noise, however, this is not a problem in our methods. As mentioned earlier, HOG features are sensitive to the edges, but these large areas usually occur along the edges, so the difference in computed features will not be significant.

\subsection{Trimming}
\label{Trimming}

In most batches we can find videos with quite long parts, at the beginning or at the end of the video, where nothing important happens. Sometimes the user is not moving at all, sometimes trying to turn on/off the recorder.\footnote{An example is batch devel$12$, video $23$.} Another problem occurring less often is in batches, where gestures are rather static. There is often variation in time the user stays in a particular gesture setting.\footnote{An example is batch devel$39$, particularly video $18$.} This is a problem for most possible approaches for tackling the one-shot-learning problem. A solution can be to remove frames from the beginning and end of the videos, as well as any part with too much inactivity.

One possible approach to removing parts of inactivity can be to watch the amount of motion in the video, and remove parts where nothing happens. This is the idea we employed.

A naive but effective way is to take the depth video and compute differences for every pixel between two consecutive frames. Taking depth videos allows us to ignore problems of texture of clothing or background. We then simply count the number of pixels whose change exceeds a given threshold, or we can simply sum the differences. After numerous experiments we ended up with Algorithm~\ref{trim}. Suppose we have a video, $n$ frames long. First we remove the background\footnote{Using an algorithm $bgremove$ provided in sample code of the Challenge \citep{ChaLearn}.} from individual frames and apply the median filter. Then we do not compute differences of consecutive frames, but rather between frames $i$ and $i+3$. This is to make the motion curve smoother and thus the method more robust. We also found that it was important to even out the amount of motion between, for instance, hand in front of body and hand in front of background. To that end, we set an upper boundary constraint on the difference at 15 (on a scale 0 to 255). Then we computed the actual motion as an average of differences between the chosen frames, as previously described, \emph{above} particular frame, for example 
\begin{align}
\label{avgmot}
motion(2) &\gets (mot(1) + mot(2)) / 2, \nonumber \\ 
motion(12) &\gets (mot(9) + mot(10) + mot(11) + mot(12)) / 4.
\end{align}
In the \emph{mot} variable we store the average change across all pixels. Then we scaled the motion to range $[0, 1]$.

\begin{algorithm}
\begin{algorithmic}
\State $n \gets length(video)$
\State $gap \gets 3$\hspace{5pt} $maxDiff \gets 15$\hspace{5pt} $threshold \gets 0.1$\hspace{5pt} $minTrim \gets 5$
\For{$i = 1 \to n$}
	\State $video(i) \gets bgremove(video(i))$ \Comment{Background removal}
	\State $video(i) \gets medfilt(video(i))$ \Comment{Median filter}
\EndFor
\For{$i = 1 \to (n-gap)$}
	\State $diff(i) \gets abs(video(i) - video(i+gap))$
	\State $diff(i) \gets min\{diff(i), maxDiff\}$
	\State $mot(i) \gets mean(diff(i))$ \Comment{Mean across all pixels}
\EndFor
\State $motion \gets avgMotion(mot)$ \Comment{As in Equation \ref{avgmot}}
\State $motion \gets scale(motion)$ \Comment{Scale motion so its range is $0$ to $1$}
\State $frames \gets vector(1:n)$
\If{$|beginSequence(motion < threshold)| \geq minTrim $}
	\State $frames \gets trimBegin(frames)$ \Comment{Remove all frames}
\EndIf
\If{$|endSequence(motion < threshold)| \geq minTrim $}
	\State $frames \gets trimEnd(frames)$ \Comment{Remove all frames}
\EndIf
\ForAll{$|sequence(motion < threshold)| > minTrim$}
	\State $frames \gets trimMiddle(sequence, frames)$ \Comment{Remove all frames but $minTrim$}
\EndFor
\State \Return $video(frames)$
\end{algorithmic}

\caption{Trimming a video}
\label{trim}
\end{algorithm}

\begin{figure}[h!]
\centering
\includegraphics[width=\linewidth]{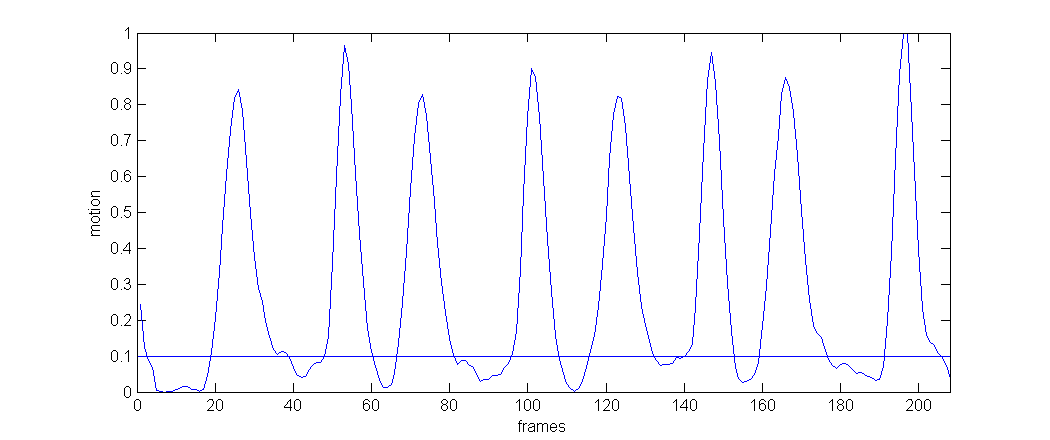}
\caption{Example of a motion graph, batch devel11, video 32}
\label{motion}
\end{figure}

Once we have the motion in the expected range, we can start actually removing frames. At first, we remove sequences from the beginning and the end of the video with motion below a $threshold$ (set to $0.1$), under the condition that they are of length at least $minTrim$ (set to $5$) frames. Then we find all sequences in the middle of the video with motion below the $threshold$ of length more than $5$, and uniformly choose $5$ frames to remain in the video. For example if we were to trim a sequence of length $13$, only frames $\{1, 4, 7, 10, 13\}$ would remain. Then we return the video with the remaining frames. Figure~\ref{motion} illustrates the threshold and the motion computed by this algorithm on a particular video.

One possible modification of this algorithm is in the step in which we scale the motion to the range of $[0, 1]$. In this case, we simply subtract $min(motion)$, and divide by $(max(motion) - min(motion))$. However, especially in videos with $4$ or $5$ gestures, sometimes large outliers cause problems, because the threshold is too big. Since the motion curve tends to be relatively smooth, instead of choosing $max(motion)$ we could choose the value of the second highest local maximum. This scaling performs slightly better on long videos, but does not work well on short videos. Since, we do not know how many gestures to expect in advance, we used the simpler method.

It is not straightforward to generalize this approach to color videos, since there is no easy way to distinguish the background from the foreground. Additionally, the texture of clothing could cause big problems to this approach. This could be overcome by adding an algorithm that would subtract the background after seeing the whole video, but we have not tried this.

\section{Feature Representation and Distance Measure}
\label{sec:feature}

In this section, we briefly describe the tools we propose for extracting features. Gestures differ from each other, both in appearance and the amount of motion while performing a particular gesture. A good descriptor of the static part of a gesture is the Histogram of Oriented Gradients, proposed by \citet{HOG}. A good method for capturing the size and direction of motion is computing the Optical Flow using the Lucas-Kanade method \citep{HOF, Lucas} and creating a histogram of flow. Motivation for these choices is explained in Section~\ref{sec:work}. Finally, we describe the Quadratic-Chi distance family proposed by \citet{Qchi} for measuring distances between histograms.

\subsection{Histogram of Oriented Gradients}

In this section we briefly describe the HOG features. The underlying idea is that the appearance and shape of a local object can often be characterized rather well by the distribution of local intensity gradient (or edge) directions, even without precise knowledge of the corresponding gradient (or edge) positions. In practice this is implemented by dividing the image window into small spatial regions (``cells''), for each cell accumulating a local 1-D histogram of gradient directions (or edge orientations) over the pixels of the cell. It is also useful to contrast-normalize the local responses before using them. This can be done by accumulating a measure of local histogram ``energy'' over larger spatial regions (``blocks'') and using the results to normalize all of the cells in the block.

\begin{figure}[h!]
\centering
\includegraphics[width=\linewidth]{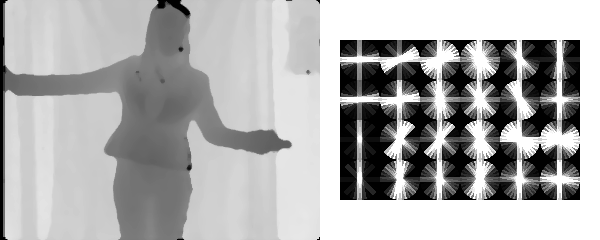}
\caption{Example visualisation of the HOG features}
\label{hog}
\end{figure}

We used a simple $[-1, 0, 1]$ gradient filter, applied in both directions and discretized the gradient orientations into $16$ orientation bins between $0^{\circ}$ and $180^{\circ}$. We had cells of size $40 \times 40$ pixels and blocks of size $80 \times 80$ pixels, each containing $4$ cells. The histogram in each cell is normalized with sum of Euclidean norms of histograms in the whole block. Each cell (except cells on the border) belongs to $4$ blocks, thus for one cell we have $4$ locally normalized histograms, the sum of which is used as the resulting histogram for the cell. Since this method cannot be used to normalize histograms of marginal cells, from $240\times 320$ image we get only $4\times 6$ spatial cells of $16$ orientation bins each. Figure~\ref{hog} provides a visual example of the HOG features at their actual resolution. The space covered is smaller than the original image, but that is not a problem, since the gestures from the dataset are not performed on the border of the frames. \citet{HOG} conclude, that fine-scale gradients, fine orientation binning, relatively coarse spatial cells, and high-quality local contrast normalization in overlapping descriptor blocks are all important for obtaining good performance.

As in Figure~\ref{hog}, we computed the HOG features from depth images, since it captures only the edges we are interested in, and not textures of clothing and so on. We used the efficient implementation from Piotr's toolbox \citep{Piotr}, function \\ \mcode{hog(image, 40, 16)}.

\subsection{Histogram of Optical Flow}

In this section we describe the general optical flow principle and the Lucas-Kanade method \citep{HOF, Lucas} for estimating the actual flow. For details we refer the reader to these works. Here we present only a brief description of the method.

The optical flow methods try to estimate the motion between two images, at times $t$ and $t + \Delta t$ at every position (in our case two consecutive frames of video). In general, the optical flow equation is formulated as a single equation with two variables. All optical flow methods introduce additional conditions for estimating the flow. The Lucas-Kanade method assumes that the flow is essentially constant in a local neighbourhood of the pixel under consideration, and solves the equation for all the pixels in the neighbourhood. The solution is obtained using the least squares principle.

After obtaining the optical flow in every point of the image we divide the image (of $240 \times 320$ pixels) to a grid of $6 \times 8$ spatial cells. We then put each optical flow vector into one of 16 orientation bins in each spatial cell, and scale them so that they sum to 1 to get a histogram of $6 \times 8 \times 16$ fields. We also tried to scale in each spatial cell separately, and the difference of error rate in our methods on all development batches was less than $0.5$. We computed the optical flow from color videos, converted to grayscale, again using efficient implementation of the Flow estimation from Piotr's toolbox \citep{Piotr}, function \\ \mcode{optFlowLk(image1, image2, [] , 4, 2, 9e-5);}

\subsection{Measuring Distance of the Histograms}
\label{subsec:distance}

\begin{figure}[h!]
\centering
\includegraphics[width=\linewidth]{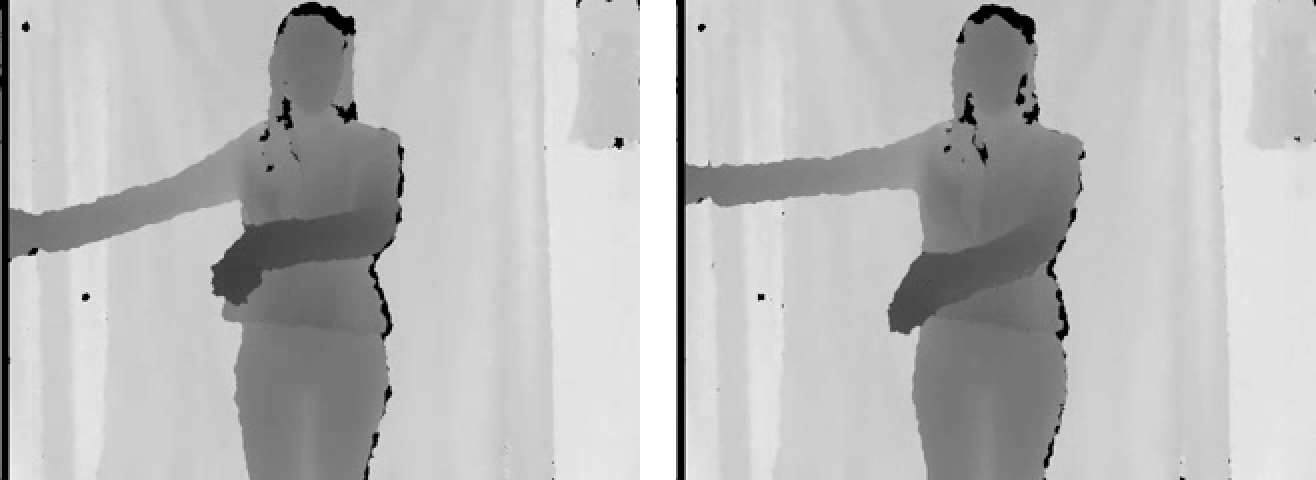}
\caption{Example of need for cross-bin similarities: the same moment in performance of the same gesture in two different videos. The right hand stays at the same place, the left hand is moving. This illustrates how the same element can result in different neighbouring orientation bins in HOG being big in different cases.}
\label{crossbin}
\end{figure}

Our method relies on making comparisons between pairs of frames in two videos, which requires as a component, to measure differences between histograms. The relatively simple methods based on the sum of bin-to-bin distances suffer from the following limitation: If the number of bins is too small, the measure is not discriminative and if it is too large it is not robust. Distances, that take into account cross-bin relationships, can be both robust and discriminative. With the HOG and HOF feature at the resolution that we selected, simple bin-to-bin comparisons are not robust, as exemplified in Figure~\ref{crossbin}. Thus we would like a measure that would look into surrounding orientation bins and, after experimenting, also to surrounding spatial cells. Thus we would also like a measure, that would reduce the effect of big differences, and also look into surrounding spatial cells. We adopted the following Quadratic-Chi distance family introduced by \cite{Qchi}.

Let $P$ and $Q$ be two histograms. Let $A$ be a non-negative symmetric bounded bin-similarity matrix, such that each diagonal element is bigger or equal to every other element in its row. Let $0 \leq m < 1$ be a normalization factor. A Quadratic-Chi histogram distance is defined as:
$$ QC_m^A\left(P, Q\right) = \sqrt{\sum_{i, j} \left(\frac{\left(P_i - Q_i\right)}{\left(\sum_c \left(P_c + Q_c\right)A_{ci}\right)^m}\right) \left(\frac{\left(P_j - Q_j\right)}{\left(\sum_c \left(P_c + Q_c\right)A_{cj}\right)^m}\right)A_{ij}}, $$
where we define $\frac{0}{0} = 0$. The normalization factor $m$ reduces the effect of big differences (the bigger it is, the bigger reduction; in our methods set to $0.5$). While comparing the $i^{th}$ orientation bins of two histograms, we want to look into the matching orientation bins, to $4$ surrounding orientation bins ($2$ left, $2$ right), and into the same orientation bins within $8$ surrounding spatial cells. MATLAB code for creating the matrix $A$ which captures these properties is in Appendix B.


\section{Recognition}
\label{sec:dtw}

In this section we describe the two methods we propose for one-shot-learning gesture recognition. We create a single model and look for the shortest path of a new video through the model in our first method. For the second method we create a separate model for every training video and using sliding frame window to look for similar parts of training videos.

\subsection{Single Model --- Dynamic Time Warping}

In this method (we will call it $SM$) we use both Histograms of Oriented Gradients and Histograms of Optical Flow and perform temporal segmentation simultaneously with recognition. 

\begin{figure}[h!]
\centering
\includegraphics[width=\linewidth]{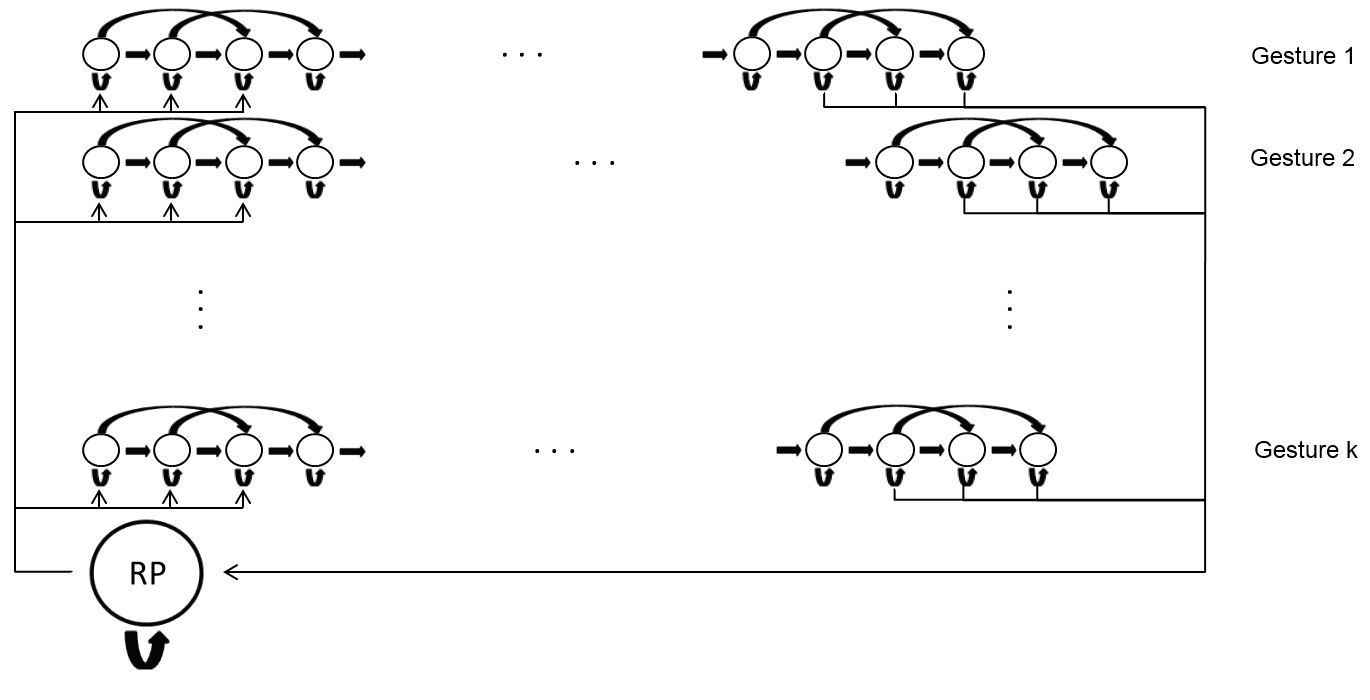}
\caption{Model for $SM$ --- Dynamic Time Warping. Each node represents a single frame of a train viedo. Each row (on the Figure) represents single train video. We add a new node --- RP, or Resting Position --- representing state where the user is not performing any particular gesture. The arrows indicate possible transitions between states (nodes).}
\label{DTW}
\end{figure}

At first, we create a model illustrated in Figure~\ref{DTW} for the whole batch in the following way: Every row in the figure represents a single training video. Every node represents a single frame of the video. In a node we store HOG and HOF features belonging to the particular frame. Recall that the HOF needs two consecutive frames. Thus if a video has $f$ frames, the representation of this video has $f-1$ nodes, ignoring the HOG of first frame. We add an arbitrary node, called Resting Position (RP), obtained as the average representation of first frames of each video.

Since we want to capture the variation in the speed of performing the gestures, we set the transitions in the following way. When being in a particular node $n$ at time $t$, moving to time $t+1$ we can either stay in the same node (slower performance), move to node $n+1$ (the same speed of performance), or move to node $n+2$ (faster performance). Experiments suggested allowing transition to node $n+3$ is not needed with the trimming described in Section~\ref{sec:preprocessing}. It even made the whole method perform worse. From the node we call RP (Resting Position) we can move to the first three nodes of any video, and from the last three nodes of every video we can move to the RP.

When we have this model, we can start inferring the gestures present in a new video. First, we compute representations of all the frames in the new video. Then we compute similarities of every node of our model with every frame representation of the new video. We compute similarities of both matching HOGs and HOFs, using the Quadratic-Chi distance described in Section \ref{subsec:distance}, and simply sum the distances. This makes sense since the empirical distribution functions of distances of HOGs and HOFs are similar. We can represent these distances as a matrix of size $N \times \left(f-1\right)$, where $N$ is the number of all nodes in the model, and $f$ is the number of frames in the new video. Using the Viterbi algorithm we find the shortest path through this matrix (we constrain the algorithm to begin in RP or in any of the first three nodes of any gesture). Every column is considered a time point, and in every time point we are in one state (row of the matrix). Between neighbouring time points the states can change only along the transitions in the model. This approach is also known as Dynamic Time Warping, \citet{DTW}.

\begin{figure}[h!]
\centering
\includegraphics[height=100mm]{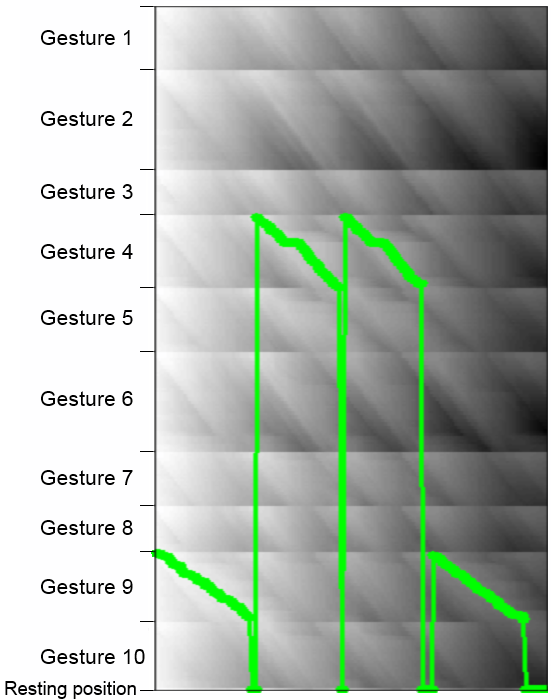}
\caption{Example of flow of states in model --- devel01, video number $11$ --- true labels are \hbox{\{9, 4, 4, 9\}}. The gray levels represent the shortest cumulative path ending in a particular point.}
\label{viterbi}
\end{figure}

The result of the Viterbi algorithm is a path, a sequence of nodes which correspond to states in which our new video was in time. From this path we can easily infer which gestures were present (which rows in Figure~\ref{DTW}), and in what order. The flow of states in time is displayed in Figure~\ref{viterbi} (the color represents the cumulative cost up to a particular point --- the darker the color, the larger the cumulative cost).

\subsection{Multiple Models --- Sliding Frame}

\begin{figure}[h!]
\centering
\includegraphics[width = \linewidth]{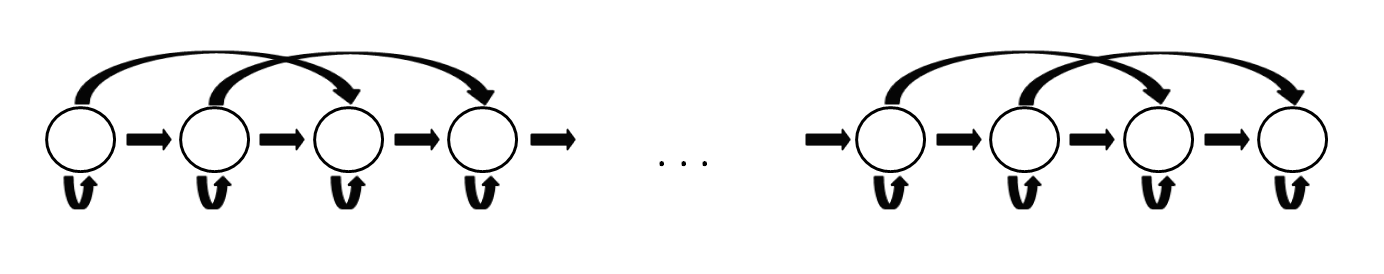}
\caption{Model for every training video in $MM$. Every node respesents a single frame of that video. The arrows indicate possible transitions between states (nodes).}
\label{slidingFrame}
\end{figure}

The second method we propose is the $MM$. Here we used only the Histogram of Oriented Gradients and perform temporal segmentation prior to recognition. We created a similar model as in $SM$, but separately for every training video, illustrated in Figure~\ref{slidingFrame}. Again, every node represents HOG features of a single frame. Thus if we have $k$ different gestures, we have $k$ similar models. We do not need an RP node, since we will be looking for short sequences in these models similar to short sequences of frames of a new video. Again, the possible transitions between states in the models, capture variation in the speed of performing gestures.

$MM$ differs from $SM$ mainly in its approach to inferring the gestures that are present. First, we compute all the HOG representations of a new video and compute their similarities with all the nodes in $k$ models. Then we employ the idea of a sliding frame. The idea is to take a small sequence of the new video and monitor the parts of the training videos that it resembles. First we select frames $1$ to $l$ (we set $l = 10$) and treat this similarly as in $SM$. We look for the shortest path through our first model without constraint on where to begin or end. We do the same with every model. This results in numerical values representing the resemblance of a small part of our new video with any part of every training video, and optionally also the number of nodes resembling it. Then we select frames $2$ to $(l+1)$, repeat the whole process, and move forward through the whole video.

\begin{figure}[h!]
\centering
\includegraphics[width = \linewidth]{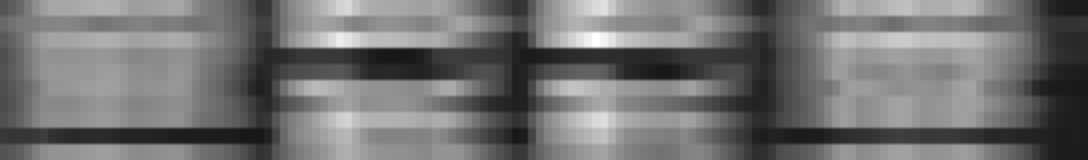}
\caption{Example of sliding frame matrix --- devel01, video number 11.}
\label{slidingScores}
\end{figure}

Finally we obtain a matrix of size $k \times (f-l+1)$, where $k$ is the number of gestures and $f$ is the number of frames in the new video. Every column represents a time instance and every row a gesture. An example of such a matrix is shown in Figure~\ref{slidingScores}. Humans can fairly easily learn to recognize where and which gestures are present, but this is a bit more challenging task for a computer. We tried to treat columns as feature vectors and feed it to $SM$ and tried to build a Hidden Markov Model to infer gestures present. We also tried to include information of what nodes of a particular model were present for every time instant, so we can prefer gestures where most of the nodes were included. That was difficult to take into account, because the start and end of most videos are very similar (Resting Position). All the methods had problems identifying two identical gestures occurring after each other, and also two similar gestures occurring after each other. We did not find satisfactory solutions to these problems without deteriorating performance.

Neither of these methods manages to beat the naive approach. We resorted to first segment the video using an algorithm provided by the organizers in the sample code called \textit{dtw\_segment}. The algorithm is very fast and segments the videos very well. After segmenting, we simply summed along the rows in corresponding parts of the scores matrix and picked the minimum. An improvement was to perform a weighed sum that emphasizes the center of the video, since the important information is usually in the middle. 

We used only HOG features in this method because every attempt to include HOF features resulted in considerably worse results. An explanation for this behaviour is we do not need to focus on the overall movement while looking only for short segments of videos, but it is more important to capture the static element. Thus the motion information is redundant in this setting.


\section{Results}
\label{sec:results}

The performance of the two methods ($SM$ \& $MM$) on the dataset is reported in this section. We also compare our results with those of other challenge participants as well as with other already published methods with experiments on this dataset. Finally we summarize our contributions and suggest an area for future work.

\subsection{Experimental results}

All our experiments were conducted on an Intel Core i7 3610QM processor, with 2 $\times$ 4GB DDR3 1600 MHz memory. The running time of $SM$ was approximately 115\% of real-time (takes longer to process than to record), while $MM$ was approximately 90\% of real-time. However, none of our methods could be trivially converted to an online method, since we need to have the whole video in advance.

The performance of our methods on all available datasets is presented in Table~\ref{results}. The results show that our preprocessing steps positively influence the final results. The $MM$ works better on the first $20$ development batches, but performs worse overall. All other published works provides results only on the first $20$ batches --- too few for any reliable conclusions. Therefore we suggest providing results on all the batches for bigger relevance.

\begin{table}[h!]
\centering
\begin{tabular}{| l | c | c |}
\hline
Batches & $SM$ & $MM$ \\
\hline
devel01-20 & 23.78 & 21.99 \\
devel01-480 & 29.40 & 34.43 \\
valid01-20 & 20.01 & 24.48 \\
final01-20 & 17.02 & 23.08 \\
final21-40 & 10.98 & 18.47 \\
\hline
\hline
devel01-20 (without trimming) & 26.24 & 22.82 \\
devel01-20 (without medfilt) & 24.70 & 23.92 \\
\hline
\hline
devel01-20 ($SM$; only HOG) & \multicolumn{2}{|c|}{24.53} \\
devel01-20 ($MM$; HOG and HOF) & \multicolumn{2}{|c|}{28.73} \\
\hline
\end{tabular}
\caption{Overview of our results on datasets. The numbers are normalized Levenshtein distances described in Section~\ref{sec:data}.}
\label{results}
\end{table}

As mentioned in Section~\ref{sec:work}, we chose our descriptors to exploit specific properties of the dataset --- the user stays at the same place, and thus the important parts of gestures occur roughly in the same position within the image. Hence it is not surprising that our model is not translation nor scale invariant. \citet{Guyon2} created 20 translated and scaled data batches, and analysed the robustness of methods of top ranking participants. In general, the bag-of-features models have this property, but they are usually rather slow. If we wanted to incorporate translation invariance, one method could be to extract body parts from the image (the algorithm is provided within Kinect Development Toolkit\footnote{\href{http://www.microsoft.com/en-us/kinectforwindows/develop/}{http://www.microsoft.com/en-us/kinectforwindows/develop/}}) and align the images so that the user is at the same position.

\begin{figure}[h!]
\includegraphics[width = \linewidth]{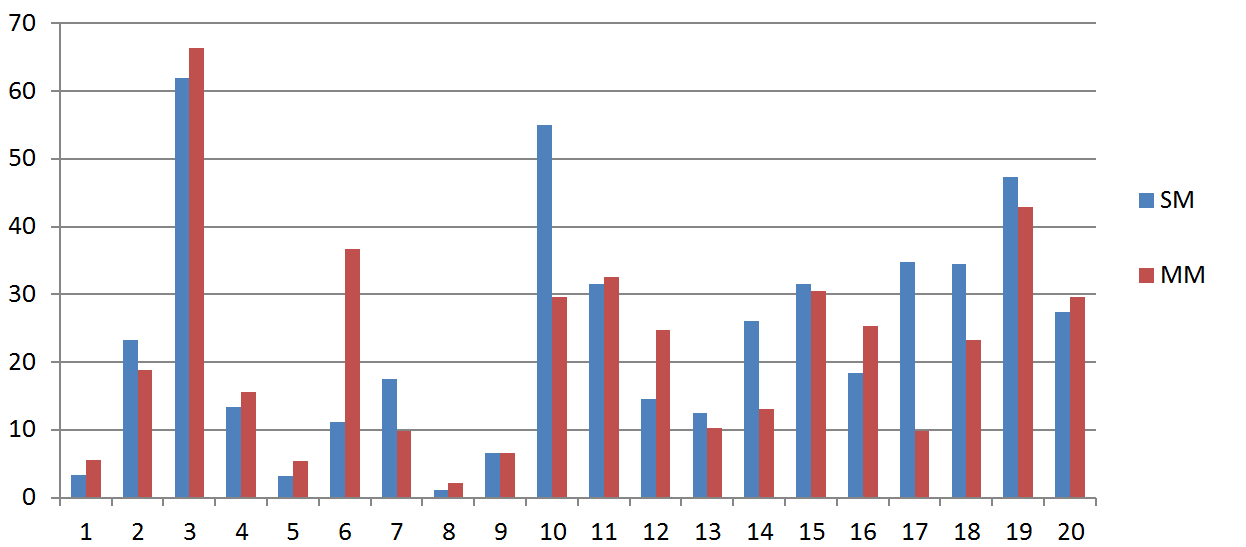}
\caption{Scores of our methods on first 20 development batches. The numbers on $y$-axis are normalized Levenshtein distances described in Section~\ref{sec:data}.}
\label{20devel}
\end{figure}

The results of our method on each of the first $20$ batches is displayed in Figure~\ref{20devel}. Often our methods perform similarly, but one can spot significant differences in batches devel06 ($SM$ --- $11.11$, $MM$ --- $36.67$), devel10 ($SM$ --- $54.95$, $MM$ --- $29.67$), devel17 \hbox{($SM$ --- $34.78$, $MM$ --- $9.78$)}. In batches devel10 and devel17, the gestures are only static and all occur in the same place in space. In this particular setting, the information about any motion (HOF) can be redundant. This could be a reason why $MM$ performs better, since we do not include any motion descriptors in the representation. In devel06, the problem is, the gestures are performed very quickly, thus the videos are often very short. This is a problem since the matrix in Figure~\ref{slidingScores} has only a few columns, resulting in poor performance of $MM$.

The above analysis brings us to a new preprocessing step. Suppose we have many algorithms for solving this one-shot-learning task. If we knew in advance which algorithm was best at recognising particular gestures, then we could boost the overall performance by selecting the `best' algorithms in advance, after seeing the training videos. This is a problem we have unsuccessfully tried to solve, and which remains open for future work. If we always pick the better from our two methods, we would achieve score of $19.04$ on the batches devel01-20.

The methods used by other challenge participants --- alfnie, Pennect, Joewan \citep{Joewan}, OneMillionMonkeys, Manavender \citep{manavender} --- are summarized by \citet{Guyon, Guyon2}. We briefly describe other published works applied on this dataset. We provide a comparison of all of these methods in Table~\ref{comparison}.

\begin{table}[h!]
\centering
\begin{tabular}{| l | c | c | c | c |}
\hline
Method / team & devel01-20 & valid01-20 & final01-20 & final21-40 \\
\hline
$SM$ (ours) & 23.78 & 20.01 & 17.02 & 10.98 \\
$MM$ (ours) & 21.99  & 24.48 & 23.08 & 18.47 \\
\hline
alfnie & NA & 9.51 & 7.34 & 7.10 \\
Pennect & NA & 17.97 & 16.52 & 12.31 \\
Joewan & 19.45 & 16.69 & 16.80 & 14.48 \\
OneMillionMonkeys & NA & 26.97 & 16.85 & 18.19 \\
Mananender & 26.34 & 23.32 & 21.64 & 19.25 \\
\hline
Wu et al. & 26.00 & 25.43 & 18.46 & 18.53 \\
BoVDW & 26.62 & NA & NA & NA \\
Lui & 28.73 & NA & NA & NA \\
Fanello et al. & 25.11 & NA & NA & NA \\
\hline
\end{tabular}
\caption{Comparison of results of methods from the competition as well as published methods. The numbers are normalized Levenshtein distances described in Section~\ref{sec:data}.}
\label{comparison}
\end{table}

\citet{Zhu} pre-segment videos and represent motions of users by Extended-Motion-History-Image and use a maximum correlation coefficient classifier. The Multi-view Spectral Embedding algorithm is used to fuse duo modalities in a physically meaningful manner.

\citet{BoVDW} present a Bag-of-Visual-and-Depth-Words (BoVDW) model for gesture recognition, that benefits from the multimodal fusion of visual and depth features. They combine HOG and HOF features with a new proposed depth descriptor.

Tensor representation of action videos is proposed by \citet{Manifolds}. The aim of his work is to demonstrate the importance of the intrinsic geometry of tensor space which yields a very discriminating structure for action recognition. The method is assessed using three gesture databases, including Chalearn gesture challenge dataset.

\citet{Fanello} develop a real-time learning and recognition system for RGB-D images. The proposed method relies on descriptors based on 3D Histogram of Flow, Global Histogram of Oriented Gradient and adaptive sparse coding. The effectiveness of sparse coding techniques to represent 3D actions is highlighted in their work.

\subsection{Contributions}

Let us now summarize our contributions. As part of the competition we managed to create solid state-of-the-art methods for the new dataset --- the goal of the competition --- which will serve as a reference point for future works. Although the crucial elements of our methods are not novel, they provide a new perspective on the possibilities of using well studied techniques, namely capturing the cross-bin relationships using the Quadratic-Chi distance. Further we present a novel algorithm for trimming videos, based only on depth data. As a preprocessing step we remove frames that bring little or no additional information, and thus make the method more robust. Experimental results show that this method does not only boost our performance, but also those of other published methods. Our detailed experiments with two very well performing methods suggest that different kinds of settings require different methods for the best performance. In particular, the possibility of choosing from more different types of models (like ours and bag-of-features) under different motion conditions remain unstudied and an open problem.


\section{Discussion and Conclusions}
\label{sec:conclusion}

In this paper we presented two methods for solving the one-shot-learning gesture recognition task introduced in the \textit{ChaLearn Gesture Challenge} \citep{ChaLearn}. We have significantly helped narrow the gap between human and machine performance (the baseline method achieved $50\%$ error rate on final evaluation set, our method $11\%$, while the human error rate is under $2\%$). Our methods outperform other published methods and we suggest that other authors provide results on the whole dataset for greater relevance of achieved results. 

We combine static --- Histograms of Oriented Gradients --- and dynamic --- Histogram of Optical Flow --- descriptors in the first method, where we create one model and perform temporal segmentation simultaneously with recognition using Dynamic Time Warping. We use only static descriptors and use pre-segmentation as a preprocessing step in the second method, where we look for similar parts in the training videos using a sliding frame.

Our first method is similar to the one developed by team Pennect in the Challenge, and also performs similarly. They also used HOG features, but at different scales, and used a one-vs-all linear classifier, while we use the Quadratic-Chi distance \citep{Qchi} to measure distances between individual frames. The recognition was also parallel with temporal segmentation using a DTW model. Surprisingly, the Pennect team used only the color images. 

Bag-of-features models provide comparable \citep{Joewan} or slightly worse \citep{BoVDW} results than ours. The advantage of these models is that they are scale and translation invariant - which is necessary for real-world applications like in gaming industry. On the other hand, these methods rely on presegmentation of videos to single gestures, and are considerably slower, hence are currently not applicable. An interesting property of these methods is their results seem to have lower variance --- error rate at difficult datasets (for instance devel10) is smaller, but struggle to obtain strong recognition rate on easy datasets (devel08, devel09).

We present a novel video trimming technique, based on the amount of motion. Its motivation is to remove unimportant segments of videos and thus reduce the probability of confusing gestures. The method improves overall results of our methods (Table~\ref{results}), and small improvement was confirmed by \citet{Zhu} --- 2\% and \citet{Joewan} --- 0.5\%.

Finally, we suggest an area for future work. Having more well working methods at our disposal, we can analyse their results on different types of gesture vocabularies, users and other settings. Overall performance could be boosted if we were able to decide which recognizer to use in advance. Especially, deeper analysis of the differences of results between Bag-of-words models and Dynamic Time Warping models is needed to obtain better description of their behaviour on different types of gesture recognition tasks.

\section*{Appendix A.}
\label{app:complexity}

In this appendix, we analyse the computational complexity of our methods.

Let us first describe the computational complexity of the building blocks of our algorithms. Let $r, c$ be the resolution of our videos. For this dataset we have $r = 240, c = 320$. Let $P$ denote number of pixels ($P = rc$). Computing both HOG and HOF features requires performing a fixed number of iterations for every pixel. Creating histograms in spatial cells requires a fixed number of operations with respect to the size of these cells. Thus the complexity of computing HOG and HOF descriptors for one example requires $\mathcal{O}(P)$ operations. Let $m$ be the number of pixels used in the median filter for every pixel. Since computing the median requires ordering, the complexity of filtering an image requires $\mathcal{O}(P m \log m)$ operations. In total, for both $SM$ and $MM$, the whole training on a batch of $N$ frames in total requires $\mathcal{O}(N P m \log (m))$ operations.

Before evaluating a new video of $F$ frames, we have to compute the representations of the frames, which is done in $\mathcal{O}(F P m \log m)$ operations. In both methods we then perform a Viterbi search. In $MM$ this is divided into several searches, but the total complexity stays the same. The most time consuming part is computing the Quadratic-Chi distances (Subsection~\ref{subsec:distance}) between all $F N$ pairs of frames from the new video and model. Computing the distance needs sum over elements over sparse $H \times H$ matrix ($H$ being the size of the histograms used) described in Algorithm~\ref{simil}. The number of non-zero elements is linear in $H$. Thus, the overall complexity of evaluating a new video is $\mathcal{O} (N P m \log(m) + N F H).$

To summarize, the running time of our methods is linear in the number of training frames, number of frames of a new video, number of pixels of a single frame, and size of histogram (number of spatial cell times number of orientation bins). Dependence on size of the filtering region for every pixel is linearithmic since it requires sorting.

\section*{Appendix B.}
\label{app:algorithm}

In this Appendix, we provide MATLAB algorithm for creating similarity matrix used in the Quadratic-Chi distance described in Section~\ref{subsec:distance}. We have histograms of $h \times w$ spatial cells, and $p$ orientation bins in each of the spatial bins. The size of the final matrix is $H \times H$, where $H = hwp$.

\begin{algorithm}[h!]
\begin{lstlisting}
gauss = fspecial('gaussian', 3, 0.56);
B = diag(ones(1,h)) + 2*(diag(ones(1, h-1), 1) + diag(ones(1, h-1), -1));
C = diag(ones(1,w)) + 2*(diag(ones(1, w-1), 1) + diag(ones(1, w-1), -1));
D = kron(C, B); % Kronecker tensor product
D(D == 1) = gauss(5); 
D(D == 2) = gauss(2); 
D(D == 4) = gauss(1);
A = imfilter( eye(p), gauss, 'circular');
A = sparse(kron(D, A)); % The final similarity matrix
\end{lstlisting}
\caption{MATLAB code producing the similarity matrix}
\label{simil}
\end{algorithm}

\vskip 0.2in
\bibliography{paper}

\begin{thebibliography}{26}
\providecommand{\natexlab}[1]{#1}
\providecommand{\url}[1]{\texttt{#1}}
\expandafter\ifx\csname urlstyle\endcsname\relax
  \providecommand{\doi}[1]{doi: #1}\else
  \providecommand{\doi}{doi: \begingroup \urlstyle{rm}\Url}\fi

\bibitem[Bay et~al.(2006)Bay, Tuytelaars, and Van~Gool]{surf}
Herbert Bay, Tinne Tuytelaars, and Luc Van~Gool.
\newblock Surf: Speeded up robust features.
\newblock In \emph{Computer Vision--ECCV 2006}, pages 404--417. Springer, 2006.

\bibitem[Berndt and Clifford(1994)]{DTW}
Donald~J. Berndt and James Clifford.
\newblock Using dynamic time warping to find patterns in time series.
\newblock In \emph{KDD workshop}, volume~10, pages 359--370, 1994.

\bibitem[ChaLearn()]{ChaLearn}
ChaLearn.
\newblock {C}ha{L}earn {G}esture {D}ataset ({CGD}2011), {C}ha{L}earn,
  {C}alifornia.
\newblock \url{http://gesture.chalearn.org/data}, 2011.

\bibitem[Chatzis et~al.(2012)Chatzis, Kosmopoulos, and Doliotis]{chatzis}
Sotirios~P Chatzis, Dimitrios~I Kosmopoulos, and Paul Doliotis.
\newblock A conditional random field-based model for joint sequence
  segmentation and classification.
\newblock \emph{Pattern recognition}, 2012.

\bibitem[Dalal and Triggs(2005)]{HOG}
Navneet Dalal and Bill Triggs.
\newblock Histograms of oriented gradients for human detection.
\newblock In \emph{Computer Vision and Pattern Recognition, 2005}, volume~1,
  pages 886--893. IEEE, 2005.

\bibitem[Doll\'ar()]{Piotr}
Piotr Doll\'ar.
\newblock {P}iotr's {I}mage and {V}ideo {M}atlab {T}oolbox ({PMT}).
\newblock \url{http://vision.ucsd.edu/~pdollar/toolbox/doc/index.html}.

\bibitem[Fanello et~al.(2013)Fanello, Gori, Metta, and Odone]{Fanello}
Sean~Ryan Fanello, Ilaria Gori, Giorgio Metta, and Francesca Odone.
\newblock One-shot learning for real-time action recognition.
\newblock 2013.

\bibitem[Guyon et~al.(2012)Guyon, Athitsos, Jangyodsuk, Hamner, and
  Escalante]{Guyon}
Isabelle Guyon, Vassilis Athitsos, Pat Jangyodsuk, Ben Hamner, and Hugo~Jair
  Escalante.
\newblock Chalearn gesture challenge: Design and first results.
\newblock In \emph{Computer Vision and Pattern Recognition Workshops (CVPRW),
  2012 IEEE Computer Society Conference on}, pages 1--6. IEEE, 2012.

\bibitem[Guyon et~al.(2013)Guyon, Athitsos, Jangyodsuk, Escalante, and
  Hamner]{Guyon2}
Isabelle Guyon, Vassilis Athitsos, Pat Jangyodsuk, Hugo~Jair Escalante, and Ben
  Hamner.
\newblock Results and analysis of the chalearn gesture challenge 2012.
\newblock 2013.

\bibitem[Hern\'andez-Vela et~al.(2012)Hern\'andez-Vela, Bautista, Perez-Sala,
  Ponce, Bar\'o, Pujol, Angulo, and Escalera]{BoVDW}
Antonio Hern\'andez-Vela, Miguel~\'Angel Bautista, Xavier Perez-Sala, Victor
  Ponce, Xavier Bar\'o, Oriol Pujol, Cecilio Angulo, and Sergio Escalera.
\newblock {B}o{VDW}: Bag-of-{V}isual-and-{D}epth-{W}ords for gesture
  recognition.
\newblock In \emph{International Conference on Pattern Recognition}, pages
  449--452, 2012.

\bibitem[Ikizler and Forsyth(2007)]{ikizler}
Nazl{\i} Ikizler and David Forsyth.
\newblock Searching video for complex activities with finite state models.
\newblock In \emph{Computer Vision and Pattern Recognition, 2007. CVPR'07. IEEE
  Conference on}, pages 1--8. IEEE, 2007.

\bibitem[Kanade and Lucas(1981)]{HOF}
Takeo Kanade and Bruce~D. Lucas.
\newblock An iterative image registration technique with an application to
  stereo vision.
\newblock In \emph{Proceedings of the 7th international joint conference on
  Artificial intelligence}, 1981.

\bibitem[Klaser and Marszalek(2008)]{klaser}
Alexander Klaser and Marcin Marszalek.
\newblock A spatio-temporal descriptor based on 3d-gradients.
\newblock 2008.

\bibitem[Laptev(2005)]{laptev}
Ivan Laptev.
\newblock On space-time interest points.
\newblock \emph{International Journal of Computer Vision}, 64\penalty0
  (2-3):\penalty0 107--123, 2005.

\bibitem[Laptev et~al.(2008)Laptev, Marszalek, Schmid, and Rozenfeld]{laptev2}
Ivan Laptev, Marcin Marszalek, Cordelia Schmid, and Benjamin Rozenfeld.
\newblock Learning realistic human actions from movies.
\newblock In \emph{Computer Vision and Pattern Recognition, 2008. CVPR 2008.
  IEEE Conference on}, pages 1--8. IEEE, 2008.

\bibitem[Lewis(1998)]{bow}
David~D Lewis.
\newblock Naive (bayes) at forty: The independence assumption in information
  retrieval.
\newblock In \emph{Machine learning: ECML-98}, pages 4--15. Springer, 1998.

\bibitem[Lowe(1999)]{lowe}
David~G Lowe.
\newblock Object recognition from local scale-invariant features.
\newblock In \emph{Computer vision, 1999. The proceedings of the seventh IEEE
  international conference on}, volume~2, pages 1150--1157. Ieee, 1999.

\bibitem[Lucas(1984)]{Lucas}
Bruce~D. Lucas.
\newblock \emph{Generalized Image Matching by the Method of Differences}.
\newblock PhD thesis, Robotics Institute, Carnegie Mellon University, July
  1984.

\bibitem[Lui(2012)]{Manifolds}
Yui~Man Lui.
\newblock Human gesture recognition on product manifolds.
\newblock \emph{Journal of Machine Learning Research}, 13:\penalty0 3297--3321,
  2012.

\bibitem[Malgireddy et~al.(2012)Malgireddy, Inwogu, and
  Govindaraju]{manavender}
Manavender~R Malgireddy, Ifeoma Inwogu, and Venu Govindaraju.
\newblock A temporal bayesian model for classifying, detecting and localizing
  activities in video sequences.
\newblock In \emph{Computer Vision and Pattern Recognition Workshops (CVPRW),
  2012 IEEE Computer Society Conference on}, pages 43--48. IEEE, 2012.

\bibitem[Pele and Werman(2010)]{Qchi}
Ofir Pele and Michael Werman.
\newblock The quadratic-chi histogram distance family.
\newblock \emph{Computer Vision--ECCV 2010}, pages 749--762, 2010.

\bibitem[Wan et~al.(2013)Wan, Ruan, Li, and Deng]{Joewan}
Jun Wan, Qiuqi Ruan, Wei Li, and Shuang Deng.
\newblock One-shot learning gesture recognition from rgb-d data using bag of
  features.
\newblock \emph{submitted to JMLR}, 2013.

\bibitem[Wang et~al.(2009)Wang, Ullah, Klaser, Laptev, Schmid, et~al.]{wang}
Heng Wang, Muhammad~Muneeb Ullah, Alexander Klaser, Ivan Laptev, Cordelia
  Schmid, et~al.
\newblock Evaluation of local spatio-temporal features for action recognition.
\newblock In \emph{BMVC 2009-British Machine Vision Conference}, 2009.

\bibitem[Wang et~al.(2006)Wang, Quattoni, Morency, Demirdjian, and
  Darrell]{wang06}
Sy~Bor Wang, Ariadna Quattoni, L-P Morency, David Demirdjian, and Trevor
  Darrell.
\newblock Hidden conditional random fields for gesture recognition.
\newblock In \emph{Computer Vision and Pattern Recognition, 2006 IEEE Computer
  Society Conference on}, volume~2, pages 1521--1527. IEEE, 2006.

\bibitem[Wu et~al.(2012)Wu, Zhu, and Shao]{Zhu}
Di~Wu, Fan Zhu, and Ling Shao.
\newblock One shot learning gesture recognition from rgbd images.
\newblock In \emph{Computer Vision and Pattern Recognition Workshops (CVPRW),
  2012 IEEE Computer Society Conference on}, pages 7--12. IEEE, 2012.

\bibitem[Xia et~al.(2010)Xia, Tao, Mei, and Zhang]{MSE}
Tian Xia, Dacheng Tao, Tao Mei, and Yongdong Zhang.
\newblock Multiview spectral embedding.
\newblock \emph{Systems, Man, and Cybernetics, Part B: Cybernetics, IEEE
  Transactions on}, 40\penalty0 (6):\penalty0 1438--1446, 2010.

\end{thebibliography}

\end{document}